\documentclass[a4paper,11pt]{article}
\usepackage[utf8]{inputenc}
\usepackage{graphicx}
\usepackage{hyperref} 
\usepackage{amssymb}
\usepackage{amsmath}
\usepackage{xcolor}
\usepackage{bbm, dsfont}
\usepackage{multicol}
\usepackage{centernot}
\usepackage{overpic}
\usepackage{subfig}

\newcommand{\p}{\mathbb{P}}
\newcommand{\R}{\mathbb{R}}

\newtheorem {Proposition}{Proposition}[section]

\newtheorem {theo}[Proposition]{Theorem}

\numberwithin{equation}{section}

\title{A survey of  bias in Machine Learning through the prism of Statistical Parity for the Adult Data Set}
\author{
  P. Besse$^1$ $\And$  E. del Barrio$^2$ $\And$  P. Gordaliza$^{1,2}$ $\And$ J-M. Loubes$^1$ $\And$ L. Risser$^1$  \\
      \\
  1\thanks{ We thank the AI interdisciplinary institute ANITI, grant agreement ANR-19-PI3A-0004 under the French investing for the future PIA3 program.} : Institut de Math\'ematiques de Toulouse\\
  INSA, Universit\'e Toulouse 3, CNRS\\
  Toulouse, France  \\
 \\
2: Instituto de Matemáticas de la Universidad de Valladolid\\
Dpto. de Estadistica e Investigacion Operativa \\
 Universidad de Valladolid\\
  Valladolid, Spain}
  
\begin{document}
	\maketitle
\begin{abstract}
	Applications based on Machine Learning models have now become an indispensable part of the everyday life and the professional world. A critical question then recently arised among the population: Do algorithmic decisions convey any type of discrimination against specific groups of population or minorities? In this paper, we show the importance of understanding how a bias can be introduced into automatic decisions. We first present a mathematical framework for the fair learning problem, specifically in the binary classification setting. We then propose to quantify the presence of bias by using the standard \textit{Disparate Impact} index on the real and well-known \textit{Adult income} data set. Finally, we check the performance of different approaches aiming to reduce the bias in binary classification outcomes. Importantly, we show that some intuitive methods are ineffective. This sheds light on the fact trying to make fair machine learning models may be a particularly challenging task, in particular when the training observations contain a bias.
\end{abstract}

\noindent%
{\it Keywords:}  Fairness, Disparate Impact,  Machine Learning, Tutorial.

\tableofcontents
\section{Introduction}
\label{sec:intro}

Fairness has become one of the most popular topics in machine learning over the last years and the research community is investing a large amount of effort in this area. The main motivation is the increasing impact that the lives of Human beings are experiencing due to the generalization of machine learning systems in a wide variety of fields. Originally designed to improve recommendation systems in the internet industry, they are now becoming an inseparable part of our daily lives since more and more companies start integrating Artifitial Intelligence (AI) into their existing practice or products. While some of these quotidian uses may involve leisure, with  vain  consequences (Amazon or Netflix use recommender systems to present a customized page that offers their products according to the order of preference of each user), other ones entail particularly sensitive decisions such as in Medicine, where patient suitability for treatment is considered; in Human Resources, where candidates are sorted out on an algorithmic decision basis; in the Automotive industry, with the release of self-driving cars; in the Banking and Insurance industry, which characterize customers according to a risk index; in Criminal justice, where the COMPAS algorithm is used in the United States for recidivism prediction... For a more detailed background on these facts see for instance \cite{romei_ruggieri_2014}, \cite{berk2018fairness} \cite{pedreschi2012study} or \cite{2018arXiv180204422F}, and references therein.

The technologies that AI offers certainly make life easier. It is however a common misconception that they are absolutely objective. In particular, machine learning algorithms which are meant to automatically take accurate and efficient decisions that mimic and even sometimes outmatch human expertise, rely heavily on potentially biased data. It is interesting to remark that this bias is often due to an inherent social bias existing in the population that is used to generate the training dataset of the machine learning models. A list of potential causes for the discriminatory behaviours that machine learning algorithms may exhibit, in the sense that groups of population are treated differently, is given in \cite{barocas2016big}. Various real and striking cases that can be found in the literature are the following. In \cite{compas}, it was found that the algorithm COMPAS used for recidivism prediction produces much higher rate  of false positive predictions for black people than for white people. Later in \cite{lahoti2019ifair}, a job platform similar to Linkedin called XING was found to predict less highly ranked qualified male candidates than female candidates. Publicly available commercial face recognition online services provided by Microsoft, Face++, and IBM respectively were also recently found to suffer from achieving much lower accuracy on females with darker skin color in \cite{buolamwini2018gender}. Although a discrimination may appear naturally and could be thought as acceptable, as in \cite{5694053} for instance, quantifying the effect of a machine learning predictor with respect to a given situation is of high importance. Therefore, the notion of fairness in machine learning algorithms has received a growing interest over the last years. We believe this is crucial in order to  guarantee a fair treatment for every subgroup of population, which will contribute to reduce the growing distrust of machine learning systems in the society.

Yet providing a definition of fairness or equity in machine learning is a complicated task and several propositions have been formulated. First described in terms of law \cite{winrow2009disparity}, fairness is now quantified in order to detect biased decisions from automatic algorithms.  We will focus on the issue of biased training data, which is one of the several possible causes of such discriminatory outcomes in machine learning mentioned above. In the fair learning literature, fairness is often defined with respect to selected variables, which are commonly denoted \textit{protected} or \textit{sensitive attributes}. We note that throughout the paper we will use both terms indistinctly. This variables encode a potential risk of discriminatory information in the population that should not be used by the algorithm. In this framework, two main streams of understanding fairness in machine learning have been considered. The probabilistic notion underlying this division is the independence between distributions. The first one gives rise to the concept of Statistical Parity, which means the independence between the protected attribute and the outcome of the decision rule. This concept is quantified using the Disparate Impact index, which is described for instance in \cite{FFMSV}. This notion was firstly considered as a tool for quantifying discrimination as the so-called $4/5^{th}$-rule by the State of California Fair Employment Practice Commission (FEPC) in 1971. For more details on the origin and first applications of this index we refer to \cite{biddle2006adverse}. The second one proposes the Equality of Odds, which considers the independence between the protected attribute and the output prediction, conditionally to the true output value. In other words, it quantifies the independence between the error of the algorithm and the protected variable. Hence, in practice, it compares the error rates of the algorithmic decisions between the different groups of the population. This second point of view has been originally proposed for recidivism of defendants in \cite{flores2016false}. Many others criteria (see for instance in \cite{berk2018fairness} for a review) have been proposed leading sometimes to incompatible formulations as stated in \cite{chouldechova2017fair}. Note finally that the notion of fairness is closely related to the notion of privacy as pointed out in \cite{dwork2012fairness}.

In this paper, our goal is to present some comprehensive statistical results  on fairness in machine learning studying the statistical parity criterion through the analysis of the example given in the \textit{Adult Income} dataset. This public dataset is available on the UCI Machine Learning Repository\footnote{\url{https://archive.ics.uci.edu/ml/datasets/adult}} and it consists in forecasting a binary variable (low or high income) which corresponds to an income lower or higher than 50k$\$$ a year. This decision could be potentially used to evaluate the credit risk of loan applicants, making this dataset particularly popular in the machine learning community. It is considered here as potentially sensitive to a discrimination with respect to the \textit{Gender} and \textit{Ethnic origin} variables. The co-variables used in the prediction as well as the true outcome are available in the dataset, hence supervised machine learning algorithms will be used. 

Section~\ref{sec:StatAnalDataset} describes this dataset. It specifically highlights the existing unbalance between the income prediction and the \textit{Gender} and \textit{Ethnic origin} sensitive variables. We note that a preprocessing step is needed in order to prepare the data for further analyses and the performed modifications are detailed in the Appendix~\ref{ssec:DataPreparation}. In Section~\ref{s:DI}, we then explain the statistical framework for the fairness problem, by particularly focusing on the binary classification setting. We follow the approach of the \textit{Statistical Parity} to quantify the fairness and we thus present the Disparate Impact as our preferred index for measuring the bias. Note that the bias is present in this dataset, so the machine learning decision rules learned  in this paper will be trained by using a biased dataset. Although, many criteria have been described in the fair learning literature, they are often used as a score without statistical control. In the cases where test procedures or confidence bounds are provided, they are obtained using a resampling scheme to get standardized Gaussian confidence intervals under a Gaussian assumption which does not correspond to the distribution of the observations. In this work, we promote the use of confidence intervals to control the risk of false discriminatory assessment. We then show in the Appendix~\ref{sec:ConfIntervals} the exact asymptotic distribution of the estimates of different fairness criteria obtained through the classical approach of the Delta method described in \cite{van1998asymptotic}. Then, Section~\ref{s:correction} is devoted to present some naive approaches that try to correct the discriminatory behaviour of machine learning algorithms or to test possible discriminations. Finally, Section~\ref{s:diff} is devoted to studying the efficiency of two easy way to incorporate fairness in machine learning algorithms: building a differentiate algorithm for each class of the population or adapting the decision of a single algorithm in a different way for each subpopulation. We then in Section~\ref{s:conclu}  present some conclusions for this work and thus provide a concrete pedagogical example for a better understanding of bias issues and  fairness treatment in machine learning. Proofs and more technical details are presented in the Appendix. Relevant code in Python to preprocess the Adult Income dataset and reproduce all the analysis and figures presented in this paper are available at the link \url{https://github.com/XAI-ANITI/StoryOfBias/blob/master/StoryOfBias.ipynb}. We also provide the French version of this Python notebook at
\url{https://github.com/wikistat/Fair-ML-4-Ethical-AI/blob/master/AdultCensus/AdultCensus-R-biasDetection.ipynb}. 

\section{Machine learning algorithms for the attribution of bank loans}\label{sec:StatAnalDataset}

One of the applications for which machine learning algorithms have already become firmly established is credit scoring. In order to minimize its risks, the banking industry uses machine learning models to detect the clients who 
are likely to deal with a credit loan. The FICO score in the US or the SCHUFA score in Germany are examples of these algorithmically determined credit rating scores, as well as those used by a number of Fintech startups, who are also basing their loan decisions entirely on algorithmic models \cite{hurley2016credit}\footnote{See, e.g., \url{https://www.kreditech.com/}.}. Yet, credit rating systems have been criticized as opaque and biased in \cite{pasquale2015black}, \cite{rothmann2014credit} or \cite{hurley2016credit}.

\begin{figure}[h]
	\centering
	\includegraphics[width=1.0\textwidth]{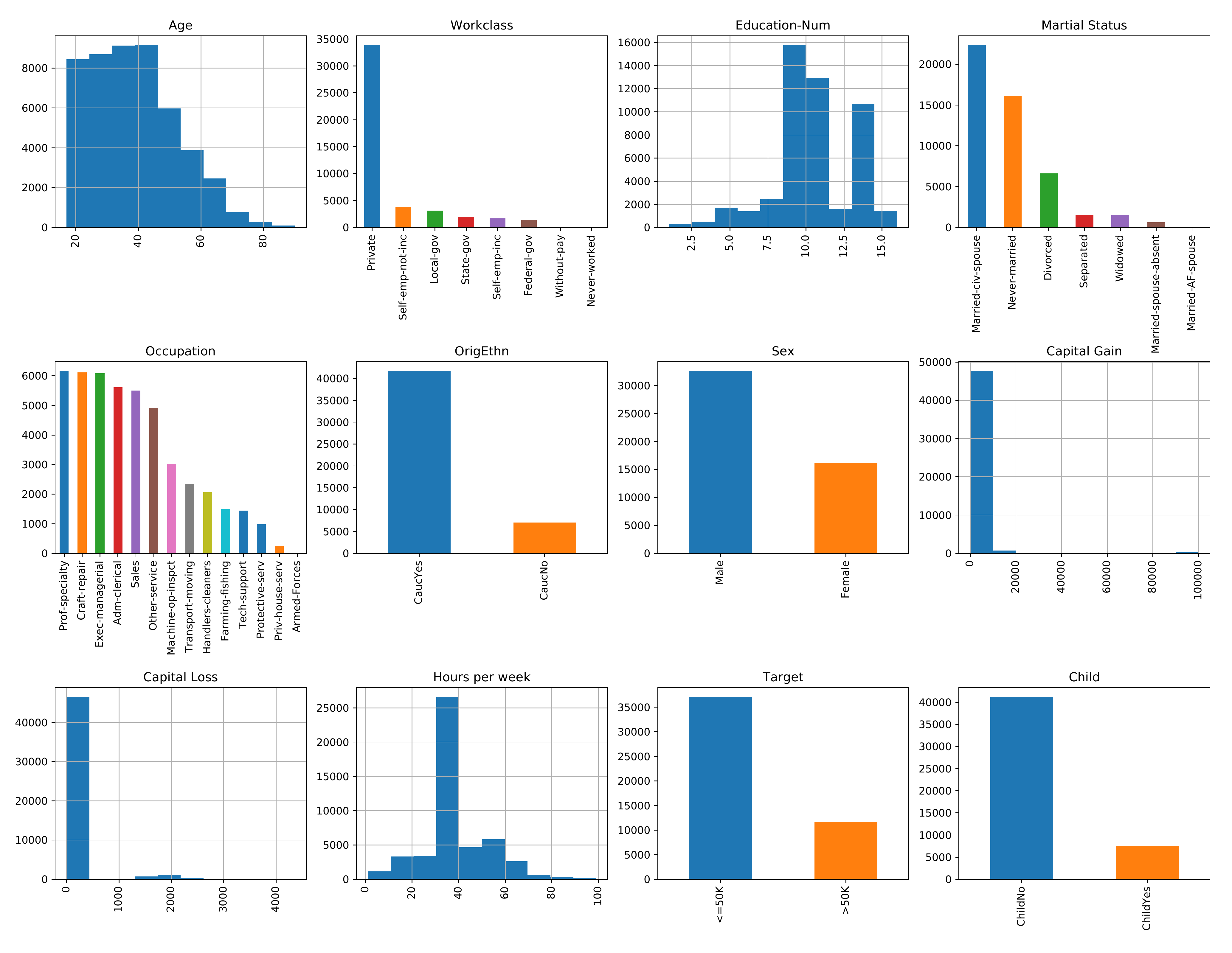}
	\caption{Adult Income dataset after pre-processing phase}
	\label{fig:AdultIncomepreprocess}
\end{figure}

In this paper, we use the Adult Income dataset as a realistic material to reproduce this kind of analyses for credit risk assessment. This dataset was built by using a database containing the results of a census made in the United States in 1994. It has been largely used among the fair learning community as a suitable benchmark to compare the performance of different machine learning methods. It contains information from about 48 thousands of individuals, each of them being described by 14 variables as detailed in Table~\ref{tab:AdultIncomeData}. This dataset is often used to predict the binary variable \textit{Anual Income higher or not than $50k \$$}. Such forecast does not convey any discrimination itself, but it illustrates what can be done in the banking or insurance industry since the machine learning procedures are similar to those  made by banks to evaluate the credit risk of their clients. The fact that the true value of the target variable is known, in contrast to the majority of the datasets available in the literature (e.g. the German Credit Data), as well as the value of potential protected attributes such as the ethnic origin or the gender, makes this dataset one of the most widely used to compare the properties of the fair learning algorithms. In this paper, we will then compare supervised machine learning methods on this dataset. A graphic representation of the distribution of each feature can be found
in~\url{https://www.valentinmihov.com/2015/04/17/adult-income-data-set/}. This representation gives a good overview of what this dataset contains. It also makes clear that it has to be pre-processed before its analysis using black-box machine learning algorithms. In this work, we have deleted missing data, errors or inconsistencies. We also have merged highly dispersed categories and eliminated strong redundancies between certain variables (see details in Supplementary material~\ref{ssec:DataPreparation}). In Figure~\ref{fig:AdultIncomepreprocess}, we represent the dataset after our pre-treatments, and show the number of occurrences for each categorical variable as well as the histograms for each continuous variable.

\subsection{Unbalanced Learning Sample}

\begin{figure}[h]
	\centering
	\includegraphics[width=0.9\textwidth]{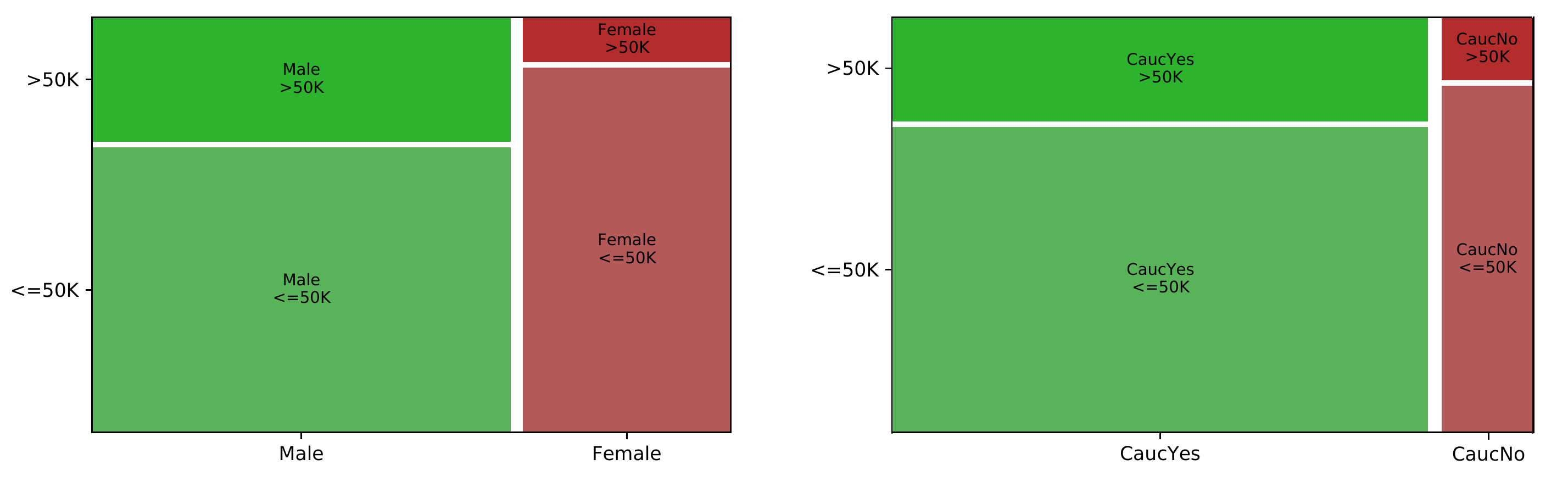}
	\caption{Enbalancement of the reference decisions in the \textit{Adult Income} dataset with respect to the \textit{Gender} and \textit{Ethnic origin} variables.}
	\label{fig:unbalancedset}
\end{figure}

After pre-processing the dataset, standard preliminary exploratory analyses first show that the dataset obviously suffers from an unbalanced repartition of low and high incomes with respect to two variables: \textit{Gender} (male or female) and \textit{Ethnic origin} (caucasian or non-caucasian). These variables therefore seem to be potentially sensitive variables in our data. Figure~\ref{fig:unbalancedset} shows this unbalanced repartition of incomes with respect to these variables. It is of high importance to be aware of such unbalanced repartitions in reference datasets since a bank willing to use an automatic algorithm to predict which clients should have successful loan applications could be tempted to train the decision rules on such unbalanced data. This fact is at the heart of our work and we question its effect on further predictions on other data. What information will be learnt from such unbalanced data: a fair relationship between the variables and the true income that will enable socially reasonable forecasts; or  biased relations in the repartition of the income with respect to the sensitive variables? We explore this question in the following section.

\subsection{Machine Learning Algorithms to forecast income}\label{ssec:BasicMLalgo}

We study now the performance of four categories of supervised learning models: logistic regression \cite{cramer2002origins}, decision trees \cite{mitchell1997machine}, gradient boosting \cite{sutton2005classification}, and Neural Network. We used the \textit{Scikit-learn} implementations of the  Logistic Regression (LR) and Decision Trees (DT), and the \textit{lightGBM} implementation of the Gradient Boosting (GB) algorithm. The Neural Network (NN) was finally coded using \textit{PyTorch} and contains four fully connected layers with Rectified Linear Units (ReLU) activation functions. 

\begin{figure}[h]
	\centering
	\includegraphics[width=1.0\textwidth]{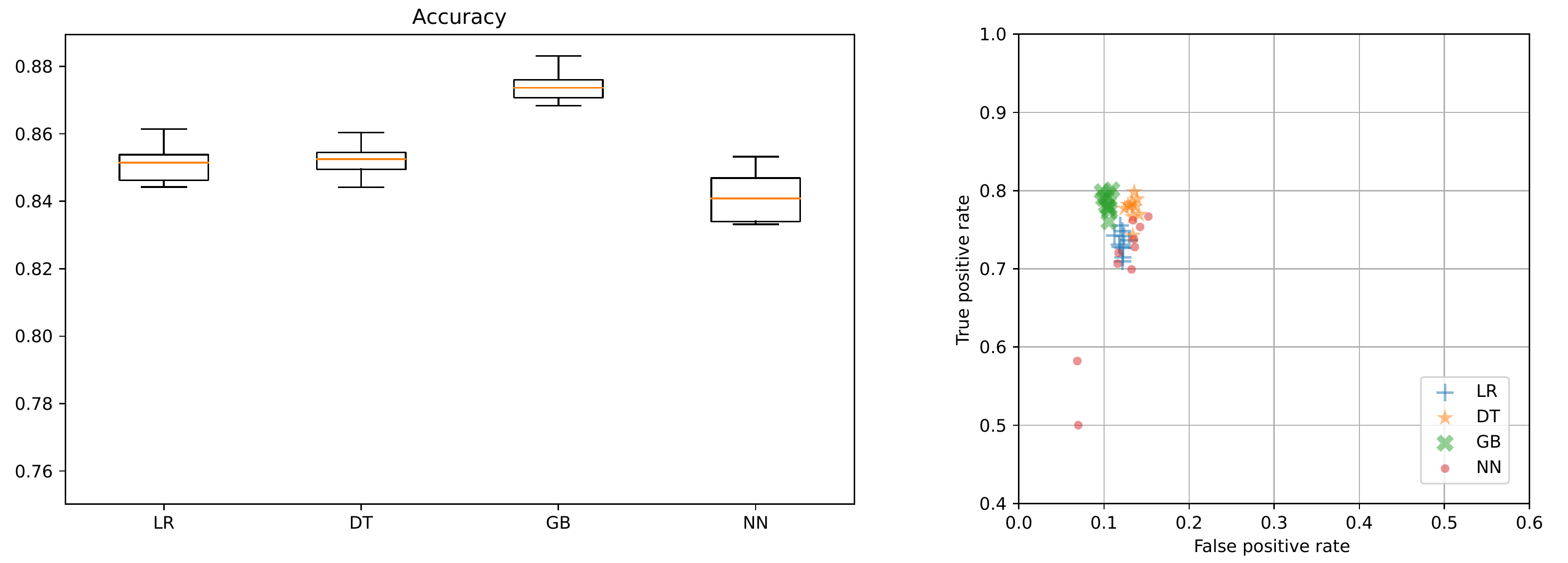}
	\caption{Prediction accuracies, true positive rates and true negative rates obtained  by using no specific treatment. Logistic Regression (LR), Decision Tree (DT), Gradient Boosting (GB) and Neural Network (NN) models were tested with 10-folds cross validation on the \textit{Adult Income} dataset.
  }
	\label{fig:MLmodels}
\end{figure}

In order to analyze categorical features using these  models, the binary categorical variables  were encoded using zeros and ones. The categorical variables with more than two classes were also transformed into one-hot vectors, \textit{i.e.} into vectors where only one element is non-zero (or hot). We specifically encoded the target variable by the values $Y=0$ for an income below $50K\$$, and $Y=1$ for an income above $50K\$$. 
We used a 10-fold cross-validation approach in order to assess the robustness of our results. 
The average accuracy as well as its true positive (TP) and true negative (TN) rates were finally measured for each trained model. Figure~\ref{fig:MLmodels} summarizes these results. 

%

We can observe in Fig.~\ref{fig:MLmodels} that the best average results are obtained by using Gradient Boosting. More interestingly, we can also remark that the prediction obtained using all models for  $Y=0$ (represented by the true negative rates) are clearly more accurate than those obtained for $Y=1$ (represented by the true positive rates), which contains about $24\%$ of the observations. 
All tested models then make more mistakes in average for the observations which should have a successful prediction than a negative one.
Note that the tested neural network is outperformed by other methods in these tests in term of prediction accuracy. Although we used default parametrizations for the Logistic Regression model as well as the Gradient Boosting model, and we simply tuned the decision tree to have a maximum depth of 5 nodes, we tested different parametrizations of the Neural Network model (number of epochs, mini-batch sizes, optimization strategies) and  kept the best performing one. It therefore appears that the neural network model we tested was clearly not adapted to the \textit{Adult Income} dataset. \\
\indent Hence we have built and compare several algorithms ranging from completely interpretable models to black box models involving optimization of several parameters. Note that we could have used the popular Random Forest algorithm that could lead to equivalent but we privilegiated boosting models whose implementation is easier using Python.



\section{Measuring the Bias with Disparate Impact}\label{s:DI}

\subsection{Notations}

Among  the criteria proposed in the literature 
to reveal the presence of a bias in a dataset or in automatic decisions (see \textit{e.g.} \cite{hardt2016equality} for a recent review), we focus in this paper on the so-called Statistical Parity. This criterion deals with the differences in reference decisions or the outcome of decision rules with respect to a sensitive attribute. Note that we only consider the binary classification problem with a single sensitive attribute for the sake of simplicity, although we could consider other tasks (\textit{e.g.} regression) or multiple sensitive attributes (see  \cite{hebert2018calibration} or \cite{kearns2018preventing}). 
Here is a summary of the notations we use:
\begin{itemize}
	\item $Y$ is the variable to be predicted. We consider here binary variables where $Y=1$ is a positive decision (here a high income) while $Y=0$ is a negative decision (here a low income);
	\item $g(X)=\hat{Y}$ is the prediction given by the algorithm. As for $Y$, this is a binary variable interpreted such that $\hat{Y}=0$  or $\hat{Y}=1$ means a negative or a positive decision, respectively. Note that most machine learning algorithms output continuous scores or probabilities. We consider in this case that this output is already thresholded.  
	\item $S$ is the variable which splits the observations into groups for which the decision rules may lead to discriminative outputs. From a legal or a moral point of view, $S$ is a sensitive variable that should not influence the decisions, but could lead to discriminative decisions. We consider hereafter that $S=0$ represents the minority that could be discriminated, while $S=1$ represents the majority. We specifically focus here on estimating the disproportionate effect with respect to two sensitive variables: the gender (male vs. female) and the ethnic origin (caucasian vs. non-caucasian). 
\end{itemize}
Statistical Parity  is often quantified in the fair learning literature using the so-called Disparate Impact (DI). The notion of DI has been introduced in the us legislation in 1971\footnote{https://www.govinfo.gov/content/pkg/CFR-2017-title29-vol4/xml/CFR-2017-title29-vol4-part1607.xml}. It measures the existing bias in a dataset as 
\begin{equation}\label{def:DIgeneral}
DI(Y,S)=\frac{\mathbb{P}(Y=1|S=0)}{\mathbb{P}(Y=1|S=1)} \,,
\end{equation}
and can be empirically estimated as
\begin{equation}\label{eq:DIestimator}
\frac{n_{10}}{(n_{00}+n_{10})} / \frac{n_{11}}{(n_{01}+n_{11})} \,,
\end{equation}
where $n_{ij}$ is number of observations such that $Y=i$ and $S=j$. 
The smaller this index, the stronger the discrimination over the minority group.
Note first that this index supposes that $\mathbb{P}(Y=1|S=0)<\mathbb{P}(Y=1|S=1)$ since $S$ is defined as the group which can be discriminated with respect to the output $Y$. 
It is also important to remark that this estimation may be unstable due to the unbalanced amount of observations in the groups $S=0$ and $S=1$ and the inherent noise existing in all data. We then propose to estimate a confidence interval around the Disparate Impact in order to provide statistical guarantees of this score, as detailed  in the Supplementary material \ref{sec:ConfIntervals}. These confidence intervals will be used later in this section to quantify how reliable are two disparate impacts computed on our dataset.
This  fairness criterion can be extended to the outcome of an algorithm by replacing in Eq.~\eqref{def:DIgeneral} the true variable $Y$ by $g(X)=\hat{Y}$, that is
\begin{equation}\label{def:DIclassifier}
DI(g,X,S)=\frac{\mathbb{P}(g(X)=1|S=0)}{\mathbb{P}(g(X)=1|S=1)}.
\end{equation} 
This measures the risk of discrimination when using the decision rules encoded in $g$ on data following the same distribution as in the test set. Hence, in \cite{gordaliza2019obtaining} a classifier $g$ is said not to have a Disparate Impact at level $\tau \in \left(0,1\right]$ when $DI(g,X,S) > \tau$. Note that the notion of DI defined Eq.~\eqref{def:DIgeneral} was first introduced as the $4/5^{th}$-rule by the State of California Fair Employment Practice Commission (FEPC) in 1971. Since then, the threshold $\tau_0=0.8$ was chosen in different trials as a legal score to judge whether the discriminations committed by an algorithm are acceptable or not (see e.g. \cite{FFMSV} \cite{ZVGG}, or \cite{mercat2016discrimination}). 

\subsection{Measures of disparate impacts}\label{ssec:MeasureDI}

The disparate impact $DI(g,X,S)$ should  be obviously close to $1$ to claim that $g$ makes fair decisions. A more subtle, though critical, remark is that it should  at least not be smaller than the general disparate impact $DI(Y,S)$. This would indeed mean that the decision rules $g$ reinforce the discriminations compared with the reference data on which it was trained.  We will then measure hereafter the disparate impacts  $DI(Y,S)$ and $DI(g,X,S)$ obtained on our dataset.



In Table~\ref{tab:OriginalBias}, we have quantified confidence intervals for the bias already present in the original dataset using Eq.~\eqref{def:DIgeneral} with the sensitive attributes \textit{Gender} and \textit{Ethnic origin}. They were computed using the method of Appendix~\ref{sec:ConfIntervals} and represent the range of values the computed disparate impacts can have with a 95\% confidence (subject to standard and reasonable hypotheses on the data). Here the DI computed on the \textit{Gender} variable then appears as very robust and the one computed on the \textit{Ethnic origin} variable is relatively robust.
It is clear from this table that both considered  sensitive attributes  generate discriminations. These discriminations are also more severe  for the \textit{Gender} variable than for the \textit{Ethnic origin} variable.
 
\begin{table}
	\centering
	\caption{Bias measured in the original dataset}
	\label{tab:OriginalBias}
	\begin{tabular}{|c|c|c|}
		\hline
		Protected attribute & DI & CI\\
		\hline
		Gender & $0.3597$ & $[0.3428,0.3765]$\\
		\hline
		Ethnic origin & $0.6006$ & $[0.5662,0.6350]$\\
		\hline
	\end{tabular}
	
\end{table}

\begin{figure}[h]
	\centering
	\includegraphics[width=1.0\textwidth]{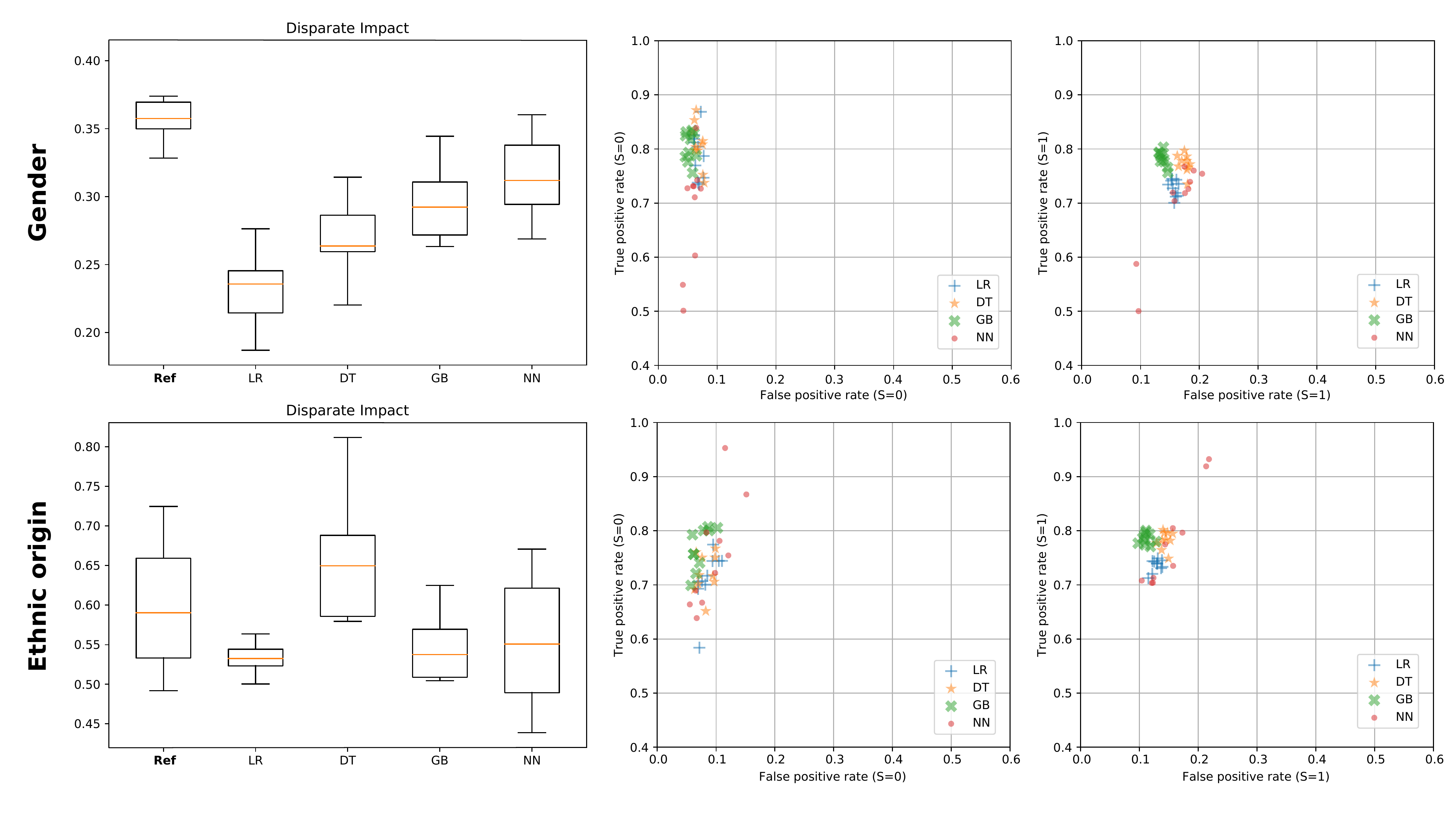}
	\caption{Bias measured in the outputs of the tested machine learning models (LR, DT, GB, NN) using the 10-folds cross validation. The disparate impacts of the reference decisions are represented by the boxplot \textit{Ref} to make clear that the unfairness is almost always re-inforced in our tests by automatic decisions. These is also a good balance between the true and the false positive decisions when the results are close to the dashed blue line.
   \textbf{(Top)} \textit{Gender} is the sensitive variable. \textbf{(Bottom)} \textit{Ethnic origin} is the sensitive variable.
  }
	\label{fig:BiasML}
\end{figure}

We have then measured the disparate impacts Eq.~\eqref{def:DIclassifier} obtained using the predictions made by the four models in the 10-folds cross-validation of Section~\ref{ssec:BasicMLalgo}. These disparate impacts are presented in Fig.~\ref{fig:BiasML}. 
We can see that, except for the decision tree with the \textit{Ethnic origin} variable, the algorithms have smaller disparate impact than for the true variable. The impact is additionally  clearly worsened with the \textit{Gender} variable using all trained predictors. These predictors therefore reinforced the discriminations in all cases by enhancing the bias present in the training sample.
Observing the true positive and true negative rates of Fig.~\ref{fig:BiasML}, which distinguish the groups $S=0$ and $S=1$ is particularly interesting here to understand this effect more deeply. 
As already mentioned Section~\ref{ssec:BasicMLalgo}, the true negative (TN) rates are generally higher than the true positive (TP) rates. 
It can be seen Fig.~\ref{fig:BiasML} that this phenomenon is clearly stronger in the subplot representing the TP and TN for $S=0$ than the one representing them for $S=1$, so false predictions are more favorable to the group $S=1$ than the group $S=0$. This explains why the disparate impacts of the predictions are higher than those of the original data (boxplots \textit{Ref} in Fig.~\ref{fig:BiasML}).
Note that these measures are directly related to the notions of equality of odds and opportunity as discussed in  \cite{hardt2016equality}.
The machine learning models we used in our experiments were then shown as unfair on this dataset, in the sense that discrimination is reinforced.

As pointed out in \cite{2018arXiv180204422F}, there may have a strong variability when computing the disparate impact of different subsamples of the data. Hence, we additionally propose in this paper an exact Central Limit Theorem to overcome this effect. The confidence intervals we obtain prove their stability when confronted to bootstrap replications and for this therefore cross-validated our results using 10 replications of different learning and test samples on the three algorithms. The construction of these confidence intervals are postponed to Section~\ref{sec:ConfIntervals} while comparison with bootstrap procedures are detailed in Section~\ref{s:boot} of the Appendix.
%
In order to conveniently compare the bias in the predictions with the one in the original data, we show on the left the bias measured in the data. We can see that these boxplots are coherent with the results of  Table~\ref{tab:OriginalBias} and Figure~\ref{fig:BiasML}, and again show that the discrimination was reinforced by the machine learning models in this test.

In all generality, we conclude here that one has to be careful when training decision rules. They can indeed worsen existing discriminations in the original database. We also remark that the majority of works  using the Disparate Impact as a measure of fairness  rely only on this score as a numerical value with no estimation of how reliable it is. This motivated the definition of our confidence intervals strategy in Appendix~\ref{sec:ConfIntervals}, which was shown to be realistic in our experiments when comparing the \textit{Ref} boxplots of Figure~\ref{fig:BiasML} with the confidence intervales of Tables~\ref{tab:OriginalBias}.
Note that we will only focus in the rest of the paper on the protected variable \textit{Gender} since it was shown in  Section~\ref{s:DI} to be clearly the variable leading to discrimination for  all tested machine learning models. We will also only test the Logistic Regression (LR) and Decision Tree (DT) as they are highly interpretable, plus the Gradient Boosting (GB) model which was shown to be the best performing one on the \textit{Adult Census} dataset.\\




\section{A quantitative evaluation of GDPR recommendations against algorithm discrimination}\label{s:correction}

Once the presence of bias is detected, the goal of machine learning becomes to reduce its impact without hampering the efficiency of the algorithm. Actually,  the predictions made by the algorithm  should remain sufficiently accurate to make the machine learning model relevant in Artificial Intelligence applications. For instance, the decisions $\hat{Y}$ made by a well balanced coin when playing \textit{head or tail} are absolutely fair, as they are independent of any possible sensitive variable $S$. However, they also do not take into account any other input information $X$, making them pointless in practice. Reducing the bias of a machine learning model $g$ therefore ideally consists in taking rid of the influence of $S$ in all input data $(X,S)$ while preserving the relevant information to predict the true outputs $Y$. We will see below that this is not that obvious, even in our simple example.

\begin{figure}[h]
	\centering
	\includegraphics[width=1.0\textwidth]{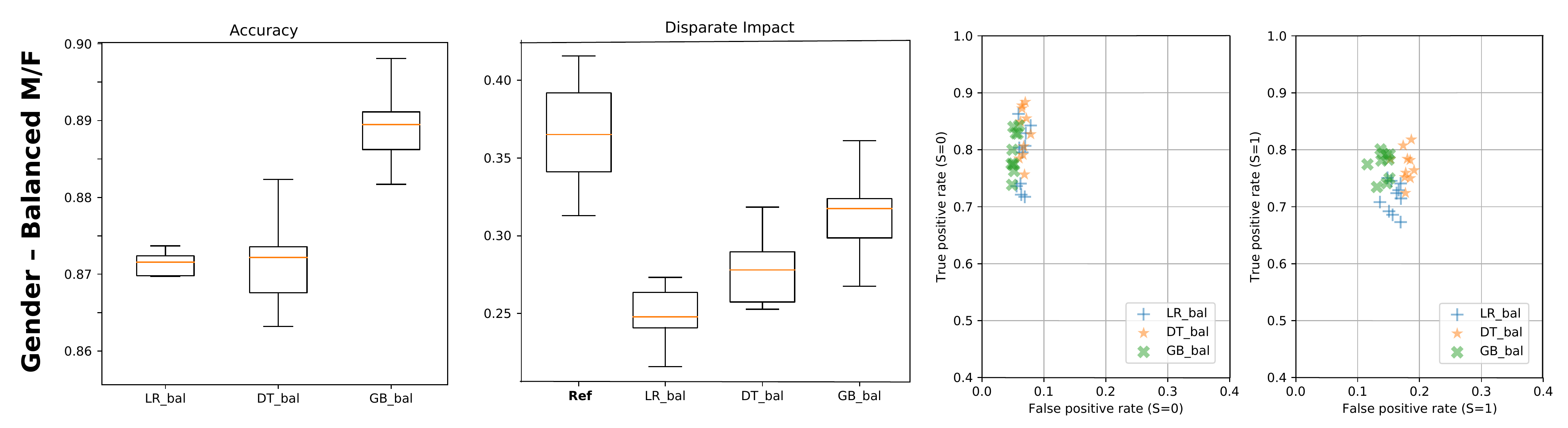}
	\caption{Bias measured in the outputs of the LR, DT and GB machine learning models using the same experimental protocol as in Section~\ref{ssec:MeasureDI} (see specifically Fig.~\ref{fig:BiasML}-\textit{(Gender)}), except that we used the same amount of males ($S=1$) and females ($S=0$) in the dataset. 
}
	\label{fig:BiasML_balanced}
\end{figure}

It is first interesting to remark that the problem cannot be solved by simply having a balanced amount of observations with $S=0$ and $S=1$. We indeed reproduced the experimental protocol of Section~\ref{ssec:MeasureDI} with 16,192 randomly chosen observations representing males (instead of 32,650), so that the decision rules were trained  in average with as many males as females. As shown in Fig.~\ref{fig:BiasML_balanced}, the trends of the results turned out to be very similar to those obtained in Fig.~\ref{fig:BiasML}-\textit{(Gender)}.

We specifically study in section  the effect of complying to the European regulations. From a legal point of view, the GDPR's recommendation indeed consists in not using the sensitive variable in machine learning  algorithms. Hence, we simply remove here $S$ from the database in subsection~\ref{ssec:sv_removed}, and we consider in subsection~\ref{ssec:testing} one of the most common   legal proof for discrimination called the {\it testing method}. It consists in considering the response for the same individual but with a different sensitive variable. We will study whether this procedure enables to detect the group  discrimination coming from the decisions of an algorithm.

\subsection{What if the sensitive variable is removed?}\label{ssec:sv_removed}

The most obvious idea to remove the influence of a sensitive variable $S$ is to remove it from the data, so we cannot use it when training the decision rules and then obviously when making new decisions. Note that this solution is recommended by GPDR regulations. To test the pertinence of this solution, we considered the algorithms analyzed in  Sections~\ref{sec:StatAnalDataset} and \ref{s:DI} and then used them without using the \textit{Gender} variable. As in Section~\ref{s:DI}, a 10-fold cross-validation approach was used to assess the robustness of our results.

\begin{figure}[h]
	\centering
	\includegraphics[width=1.0\textwidth]{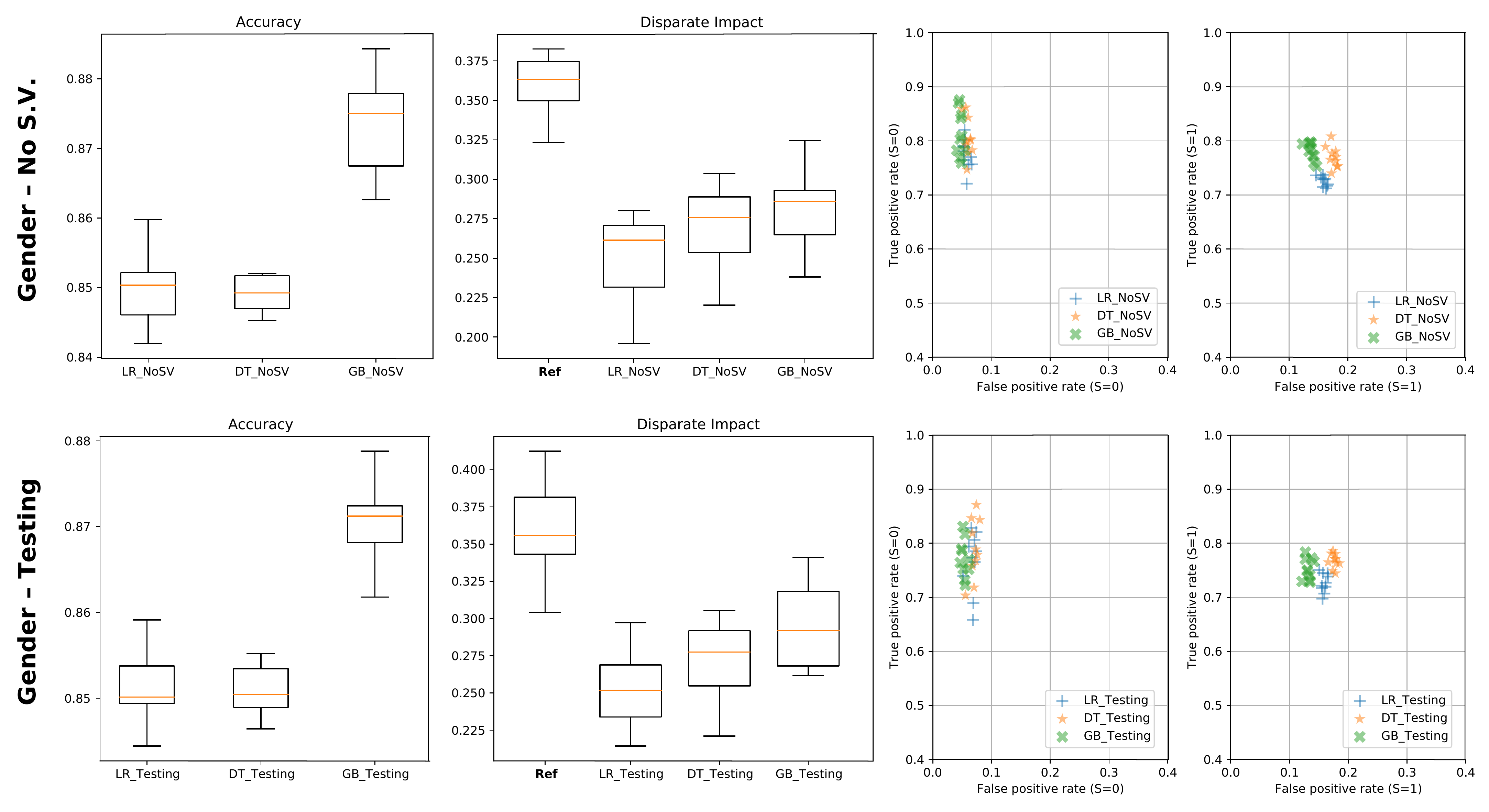}
	\caption{Performance of the machine learning models LR, DT and GB when \textbf{(top)} removing the \textit{Gender} variable, and  \textbf{(bottom)} when using a testing procedure.
  }
	\label{fig:NoGenderML_TestingGender}
\end{figure}

As shown Figure~\ref{fig:NoGenderML_TestingGender}-(top), the disparate impacts as well as the model accuracies remained almost unchanged when removing the \textit{Gender} variable from the input data. 
Anonymizing  database by removing a variable therefore had very little effect on the discrimination that is induced by the use of an automated decision algorithm. This is very likely to be explained by the fact that a machine learning algorithm 
uses all possible information conveyed by the variables. 
In particular, if  the sensitive variable (here the \textit{Gender} variable) is strongly correlated to other variables, then the algorithm learns and reconstruct automatically the sensitive variable from the other variables. Hence we can deduce that social determinism is stronger than the presence of the sensitive variable here, so the classification algorithms were not impacted by the removal of this variable. 

Obtaining fairness is a far more complicated task than this simple trick. It is at the heart of modern research on fair learning.  More complex fairness mathematical methods to reduce disparate treatment are discussed for instance in  \cite{kleinberg2016inherent} or in \cite{gordaliza2019obtaining}. 

\subsection{From Testing for bias detection to unfair prediction}\label{ssec:testing}

Testing procedures are often used as a legal proof for discrimination. For an individual  prediction, such procedures consist in first creating an artificial individual which shares the same characteristics of a chosen individual that suspects a disparate treatment and discrimination, but has a different protected variable.  Then it amounts to testing whether this artificial individual has the same prediction as the original one. If the predictions differ, then this conclusion can serve as a legal proof for discrimination. \\
\indent  These procedures have existed for a long time (since their introduction in 1939 \footnote{\url{https://fr.wikipedia.org/wiki/Test_de_discrimination}}) , and since 2006 when the French justice has taken them as a proof of biased treatment, although the testing process itself has been qualified as unfair\footnote{\url{https://www.juritravail.com/discrimination-physique/embauche/ph-alternative-A-1.html}}. Furthermore, this technique has been generalized by sociologists ans economists (see for instance \cite{riach2002field} for a description of such method) to statistically measure group discrimination in housing and labour market by conducting carefully controlled field experiments.

This testing procedure considered as a discrimination test is nowadays a commonly used method in France to assess fairness for sociological studies of \textit{Observatoire des discriminations}\footnote{\url{https://www.observatoiredesdiscriminations.fr/testing}} and  \textit{laboratoire TEPP} as pointed out in \cite{ldiscriminations}, or governemental studies DARES\footnote{\url{https://dares.travail-emploi.gouv.fr/dares-etudes-et-statistiques/etudes-et-syntheses/dares-analyses-dares-indicateurs-dares-resultats/testing}} of French Ministry of Work ISM Corum \footnote{\url{http://www.ismcorum.org/}}. Some industries are labeled using such test. An audit quality of recruiting methods is proposed while \textit{Novethic}\footnote{\url{https://www.novethic.fr/lexique/detail/testing.html}} proposes ethic formations.

Testing is efficient to detect human discrimination specially in labour market but hiring tech is producing more and more softwares or web platforms performing predictive recruitment as in \cite{fat2020}. Does testing remains valid in front of machine learning algorithms? This last strategy is evaluated using the same experimental protocol as in the previous sections. The results of these experiments are shown in Figure 6-(bottom). Testing does not detect any discrimination when the sensitive variable is captured by the other variables.

An algorithmic solution to bypass this testing procedure is given by the following trick. Train a classifier as usual using all available information $X,S$ and then build a {\it testing compliant} version of it as follows : for an individual, the predicted outcome is assigned as the  best decision obtained on the actual individual $f(x,s)$ and a virtual individual with exactly the same characteristics as the original one, except for the protected variable $s$  which has the opposite label $s^{'}$ (\textit{e.g.} the \textit{Gender} variable is \textit{Male} instead of \textit{Female}), namely $f(x,s^{'})$. Note that in case of multi-class labels, the outcome should be the most favourable decision for all possible labels. This classifier is fair by design in the sense that no matter their gender, the  {\it testing procedure} can not detect a change in the individual prediction.

Nevertheless, this trick against testing cannot cheat usual evaluation of discrimination by using a disparate impact measure which is usual in the USA by measuring the impact on real and not fictitious recruitment. This is the reason why hiring tech companies add some facilities (\cite{fat2020}) to mitigate ethnic bias of algorithmic hiring for avoiding an enterprise juridical complications.
The evaluation of this strategy is evaluated using the same experimental protocol as in the previous sections and these  are shown in Figure~\ref{fig:NoGenderML_TestingGender}-(bottom).

%
As expected for previous results,  this method has little impact on the classification errors and the disparate impacts. This emphasises  the conclusion of Section~\ref{ssec:sv_removed} claiming that the \textit{Gender} variable is captured by other variables. Removing the effect of a sensitive variable can therefore require more advanced treatments than those described above. 


\section{Differential treatment for fair decision rules} \label{s:diff}
 
\subsection{Strategies}
As we have seen previously, bias may induce discrimination of an automatic decision rule. Although many complex methods  have been developed to tackle this problem, we investigate  in this section the effects of two easy and maybe naive modifications of machine learning algorithms. We  present in this section the effect of two alternative strategies to build fair classifiers. They have in common the idea of considering different treatments according to  each group $S=\{0,1\}$.  These strategies are the following :
\begin{enumerate}
	\item \textbf{Building a different classifier for each class of the sensitive variable:} This strategy consists in training the same prediction model with different parameters for each class of the sensitive variable. We denote {\it separate treatment} this strategy.
	\item \textbf{Using a specific threshold for each class of the sensitive variable:} Here, a single classifier is trained for all data to produce a score. The binary prediction is however get using a specific threshold for each sub-group $S=0$ or $S=1$. Note that when the score is obtained by estimating the conditional distribution $\eta(x)=P(Y=1|X=x)$ then the threshold used is often $0.5$.  Here this threshold is made $S$-dependent and is adapted to avoid any possible discrimination. In practice, we keep a threshold of $0.5$ for the observations in the group $S=1$ but we adapt the corresponding threshold for the observations in the group $S=0$. In our tests, we automatically set this threshold on the training set so that the disparate impact  is close to $0.8$ in the cases where it was originally lower to this this socially accepted threshold. The classifier and the potentially adapted threshold are then used for further predictions. This corresponds in a certain way to favour  the minority class by changing equality to equity. We denote  this strategy as {\it positive discrimination} since this procedure corresponds to this purpose.
\end{enumerate}

\subsection{Results obtained using the \textit{Separate Treatment} strategy}


\begin{figure}[h]
	\centering
	\includegraphics[width=1.0\textwidth]{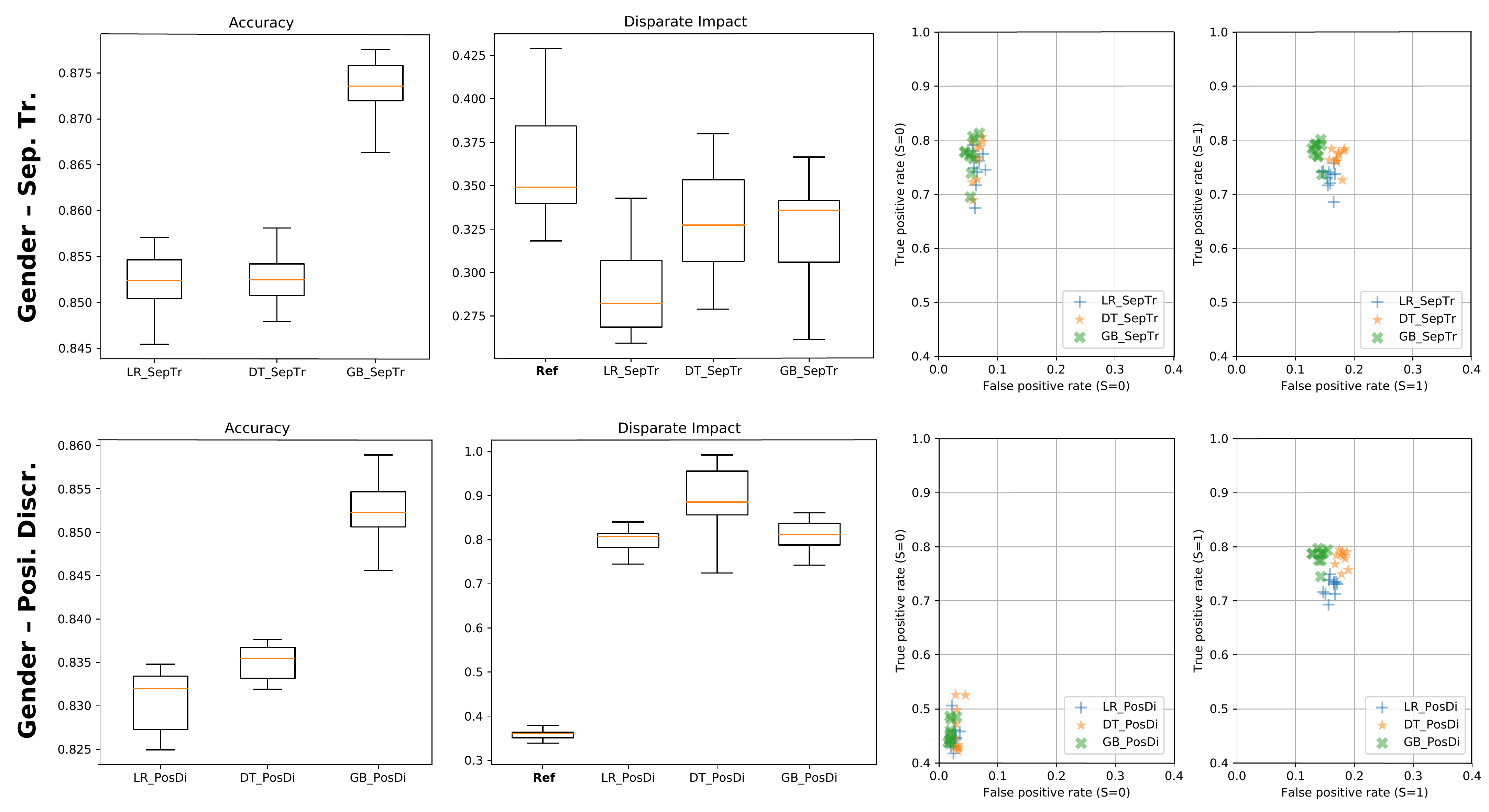}
	\caption{Performance of the machine learning models LR, DT and GB when \textbf{(top)} using a \textit{Separate Treatment} for the groups $S=0$ and $S=1$, and \textbf{(bottom)} when using a \textit{Positive Discrimination} strategy for the groups $S=0$.
  }
	\label{fig:GenderST}
\end{figure}

Splitting the model parameters into parameters adapted to each group reduces the bias of the predictions when compared to the initial model, but it does not remove it.  As we can see in Figure~\ref{fig:GenderST}-\textit{(top)}, where the notations are analogous to those in the above figures, it improved the disparate impact in all cases for relatively stable prediction accuracies. Note that the improvements are more spectacular for the basic Logistic Regression and Decision Tree models than for the Gradient Boosting model. This last model is indeed particularly efficient to capture fine high order relations between the variables, which gives less influence to the strong non-linearity generated when splitting the machine learning model into two class-specific models. Hence building different models reduces but does not solve the problem, the level of discrimination in the decisions being only slightly  closer to the level of bias in the initial dataset. 

\subsection{Results obtained using the \textit{Positive Discrimination} strategy}


Results obtained using the \textit{positive discrimination} strategy are shown in Figure~\ref{fig:GenderST}-\textit{(bottom)}. They clearly emphasize the spectacular effect of this strategy on the disparate impacts, which can be controlled by the data scientist. By adjusting the threshold, it is possible to adjust the levels of discriminations in the dataset, as in this example where the socially acceptable level of 0.8 can be reached.  In this case we see a decrease in the performance of the classifier, but  yet being reasonable.

These results should however be tempered for a main reason.  Although the average error receives little changes, the number of false positive cases of women is clearly increased when introducing positive discrimination.  In our tests more than half of the predictions that should have been false in the group $S=0$ are even true.
These false positive decisions have a limited impact on the average prediction accuracy as they where obtained in the group $S=0$ which has less observations than $S=1$ and that there are clearly less true predictions with $Y=1$ than $Y=0$. Yet false positive errors are considered as the most important error type and thus this increase may be very harmful for the decision maker. On a legal point of view, this procedure may be judged as unfair or rises political issues that are far beyond the scope of this paper.

\section{Conclusions} \label{s:conclu}

\begin{figure}[h]
	\centering
	\includegraphics[width=0.8\textwidth]{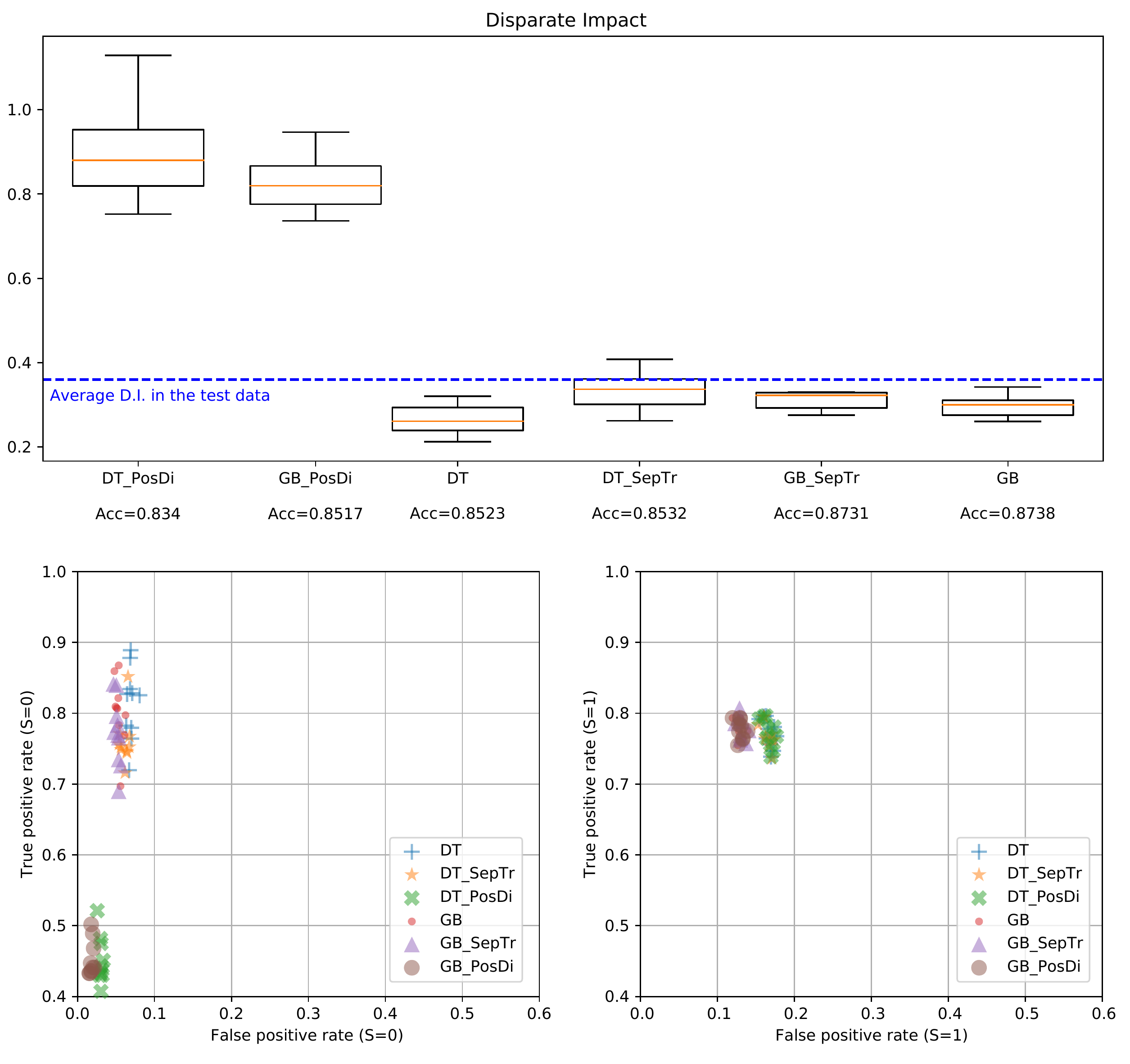}
	\caption{
    Summary of the main results: The best performing algorithms of Sections~\ref{s:DI} and \ref{s:diff} are compared here. \textbf{(top)} Boxplots of the disparate impacts from the least accurate method on the left, to the most accurate method on the right, and  \textbf{(bottom)} corresponding true positive and true negative rates in the groups $S=0$ and $S=1$.
  }
	\label{fig:summary}
\end{figure}

In this paper, we provided a case-study of the use of machine learning technics for the prediction of the well-known {\it Adult Income} dataset. We focused on a specific fairness criterion, the statistical parity, which is measured through the Disparate Impact. This metric quantifies  the difference of the behaviour of a classification rule applied for two subgroups of the population, the minority and the majority. Fairness is achieved when the algorithm behaves in the same way for both groups, hence when the sensitive variable does not play a significant role in the prediction. Main results are summarized in Figure~\ref{fig:summary}.

In particular, we convey the following take-home messages: \textbf{(1)} Bias in the training data may lead to machine learning algorithms taking unfair decisions, but not always. While there is a clear increase of bias using the tested machine learning algorithms with respect to the \textit{Gender} variable, the \textit{Ethnic Origin} does not lead to a severe bias. \textbf{(2)} As always in Statistics, computing a mere measure is not enough but confidence intervals are needed to determine the variability of such indexes. Hence, we proposed an ad-hoc construction of confidence intervals for the Disparate Impact. \textbf{(3)} Standard regulations that promote either the removal of the sensitive variable or the use of testing technics appeared as irrelevant when dealing with fairness of machine learning algorithms.

 \indent  Note also that different notions of fairness (local and global) are at stake here. We first point out that testing methods focus on individual fairness while statistical methods such as the Disparate Impact Analysis tackle the issue of group fairness.  These two notions if related to the similar notion of discrimination with respect to an algorithmic decision are yet different. In this work, we showed that an algorithm can be designed to be individually fair while still presenting a strong discrimination with respect to the minority group. This is mainly due to the fact that testing methods are unable to detect the discrimination hidden in the algorithmic decisions that are due to the training on an unbalanced sample. Testing methods detect discrimination if individuals with the same characteristics but different sensitive variables are treated in a different way. This corresponds to trying to find counterfactual explanation to an individual with a different sensitive variable. This notion of counterfactual explanations to detect unfairness has been developed in \cite{kusner2017counterfactual}. Yet the testing method fails in finding a counterfactual individual since it is not enough to change only the sensitive variable but a good candidate should be the closest individual with a different sensitive variable but with the variables that evolve depending on $S$. For this, following some recent work on fairness with optimal transport theory as in \cite{gordaliza2019obtaining} developing an idea from \cite{FFMSV}, some authors propose a new way of testing discrimination by computing such new counterfactual models in \cite{10.1145/3351095.3372845}. Finally, we tested two a priori naive solutions consisting either in building different models for each group or in choosing different rules for each group. Only the latter that can be considered as \textit{positive discrimination} proves helpful in obtaining a fair classification. Note that if some errors are increased (false positive rate), this method has a good generalization error. Yet in other cases, the loss of efficiency could be greater and this method may lead to unfair treatment.
 
 This data set has been extensively studied in the literature on fairness in machine learning and we are well aware of the numerous solutions that have been proposed to solve this issue. Even with standard methods, it is possible for a data scientist,  when confronted to fairness in machine learning, to design algorithms that have very different behaviors and yet achieving a good classification error rate. Some algorithms hamper discrimination in the society while others just maintain its level, and some others correct this discrimination and provide gender equity. It is worth noting that the most explainable algorithms, such as the logistic regression, do not protect from discrimination. On the contrary, the capture of gender bias is inmediate due to its simplicity, while more complex algorithms might be more protected from this spurious correlation or, since the variable is discrete, better said spurious dependency.
 
 The choice of a model should not be driven only by its performance with respect to a generalization error but should also be explainable in terms of bias propagation. For this, measures of fairness should be included in the evaluation of the model. In this work, we only considered statistical parity type fairness but many other definitions are available, without any consensus on the better choice for such a definition neither from a mathematical or a legal point of view. A strong research effort in data science is hence  the key for a better use of Artificial Intelligence type algorithms. This will allow data scientists to describe precisely the algorithmic designing process, as well as their behaviour, in terms of precision and propagation of bias.
 
In closing, note that biases are what enables machine learning algorithms to work and helpfulness of complex algorithms is due to their ability to find hidden bias and correlations in very large data sets. Hence bias removal should be handled with care because one part of this information is crucial, while the other is harmful. Therefore, explainability should not be understood in terms of explainability of the whole algorithm, but maybe one line of future research in machine learning should focus on explainability of the inner bias of an algorithm, or its explainability with respect to some legal regulations.


\appendix
\section{Appendix}


\subsection{The Adult Income dataset}
\begin{table}
	\caption{The Adult Income dataset}
	\label{tab:AdultIncomeData}
	\vspace{-0.4cm}
	\centering
	\begin{tabular}{|c|l|p{12cm}|}
		\hline
		Nº & Label & Possible values\\
		\hline
		1 & \texttt{Age} & Real\\
		\hline
		2 & \texttt{workClass} &
		Private, Self-emp-not-inc, Self-emp-inc,
		Federal-gov, Local-gov, State-gov, Withoutpay,
		Never-worked\\
		\hline
		3 & \texttt{fnlwgt} & Real \\
		\hline
		4 & \texttt{education} & Bachelors, Some-college, 11th, HS-grad,
		Prof-school, Assoc-acdm, Assoc-voc, 9th,
		7th-8th, 12th, Masters, 1st-4th, 10th,
		Doctorate, 5th-6th, Preschool\\
		\hline
		5 & \texttt{educNum} & integer\\
		\hline
		6 & \texttt{mariStat} & Married-civ-spouse, Divorced, Nevermarried,
		Separated, Widowed, Marriedspouse-
		absent, Married-AF-spouse\\
		\hline
		7 & \texttt{occup} & Tech-support, Craft-repair, Other-service,
		Sales, Exec-managerial, Prof-specialty,
		Handlers-cleaners, Machine-op-inspct,
		Adm-clerical, Farming-fishing, Transportmoving,
		Priv-house-serv, Protective-serv,
		Armed-Forces\\
		\hline
		8 & \texttt{relationship} & Wife, Own-child, Husband, Not-in-family,
		Other-relative, Unmarried \\
		\hline
		9 & \texttt{origEthn} & White, Asian-Pac-Islander, Amer-Indian-
		Eskimo, Other, Black\\
		\hline
		10 & \texttt{gender} & Female, Male \\
		\hline
		11 & \texttt{capitalGain} & Real \\
		\hline
		12 & \texttt{capitalLoss} & Real \\
		\hline
		13 & \texttt{hoursWeek} & Real \\
		\hline
		14 & \texttt{nativCountry} & United-States, Cambodia, England,
		Puerto-Rico, Canada, Germany, Outlying-
		US(Guam-USVI-etc), India, Japan,
		Greece, South, China, Cuba, Iran,
		Honduras, Philippines, Italy, Poland,
		Jamaica, Vietnam, Mexico, Portugal,
		Ireland, France, Dominican-Republic, Laos, Ecuador, Taiwan, Haiti, Columbia,
		Hungary, Guatemala, Nicaragua, Scotland,
		Thailand, Yugoslavia, El-Salvador,
		Trinidad and Tobago, Peru, Hong, Holand-
		Netherlands \\
		\hline
		15 & \texttt{income} & $>50k$, $\leq 50k$\\
		\hline
	\end{tabular}
	
\end{table}

\subsubsection{Data preparation}\label{ssec:DataPreparation}

As discussed in the introduction of Section~\ref{sec:StatAnalDataset}, the study has started with a detailed preprocessing of the raw data to give a more clear interpretation to further analyses. 
First, we noticed that the variable \texttt{fnlwgt} (Final sampling weight) has not a very clear meaning so it has been removed. For a complete description of such variable access the link \url{http://web.cs.wpi.edu/~cs4341/C00/Projects/fnlwgt}. We have also performed a basic and multidimensional exploration (MFCA) in order to represent the possible sources of bias in the data in  \url{https://github.com/wikistat/Fair-ML-4-Ethical-AI/blob/master/AdultCensus/AdultCensus-R-biasDetection.ipynb}.

This exploration leaded to a deep cleaning of the data set and highlighted difficulties present on certain variables, raising the need to transform some of them before fitting any statistical model. In particular, we have deleted missing data, errors or inconsistencies; grouped together certain highly dispersed categories and eliminated strong redundancies between certain variables. This phase is notoriously different from the strategy followed by \cite{2018arXiv180204422F} who analyze raw data directly. Some of these main changes are listed below:
\begin{itemize}
	\item Variable 3 \texttt{fnlwgt} is removed since it has little significance for this analysis.
	\item The binary variable \texttt{child} is created to indicate the presence or absence of children.
	\item Variable 8 \texttt{relationship} is removed since it is redundant with \texttt{gender} and \texttt{mariStat}.
	\item Variable 14 \texttt{nativCountry} is removed since it is redundant with variable \texttt{origEthn}.
	\item Variable 9 \texttt{origEthn} is transformed into a binary variable: CaucYes vs. CaucNo.
	\item Varible 4 \texttt{education} is removed as redundant with variable \texttt{educNum}.
	\item Additionally clean-up the $<50K$, $\leq 50K$, $>50K$ and $\geq 50K$ in variable ``Target"
\end{itemize} 
%
%
\subsection{Testing lack of fairness and confidence intervals}\label{sec:ConfIntervals}
Let $\left(X_i, S_i, \hat{Y}_i=g(X_i)\right), i=1, \ldots, n,$ be a random sample of independent and equally distributed variables. Previous criterion can be consistently estimated by their empirical version. Yet the value of the criterion may depend on the data sample. Due to the importance of obtaining an accurate proof of  unfairness in a decision rule it is important to obtain confidence intervals in order to control the error of detecting unfairness. In the  literature it is often achieved by computing the mean over several sampling of the data.   We provide in the following the exact asymptotic behaviors of the estimates in order to build confidence intervals.

\begin{theo}[Asymptotic behavior of the Disparate Impact estimator]
	\label{th:DI}
	Set the empirical estimator of DI(g) as
	\begin{equation*}
	T_n := \displaystyle\frac{\sum_{i=1}^{n}\mathbbm{1}_{g(X_i)=1}\mathbbm{1}_{S_i=0}\sum_{i=1}^{n}\mathbbm{1}_{S_i=1}}{\sum_{i=1}^{n}\mathbbm{1}_{g(X_i)=1}\mathbbm{1}_{S_i=1}\sum_{i=1}^{n}\mathbbm{1}_{S_i=0}}.
	\end{equation*} Then the asymptotic distribution of this quantity is given by 
	\begin{equation}\label{eq:convergenceTn}
	\frac{\sqrt{n}}{\sigma}\left(T_n-DI(g,X,S)\right) \xrightarrow{d} N(0,1),  \ as \ n\rightarrow \infty,
	\end{equation}
	where $\sigma = \displaystyle\sqrt{\nabla\varphi^T\left(\mathbb{E}Z_1\right) \Sigma_4 \nabla\varphi\left(\mathbb{E}Z_1\right)}$ and 
	\begin{equation*}
	\nabla\varphi^T\left(\mathbb{E}Z_1\right)=\left(\frac{\pi_1}{p_1\pi_0},-\frac{p_0\pi_1}{p_1^2\pi_0}, -\frac{p_0\pi_1}{p_1\pi_0^2},\frac{p_0}{p_1\pi_0} \right)
	\end{equation*}
	\begin{equation*}
	\Sigma_4= \left(\begin{array}{cccc}
	p_0(1-p_0) & & & \\
	-p_0p_1 & p_1(1-p_1) & & \\
	\pi_1p_0 & -\pi_0p_1 & \pi_0\pi_1 & \\
	-\pi_1p_0 & \pi_0p_1 & -\pi_0\pi_1 & \pi_0\pi_1
	\end{array} \right),
	\end{equation*}
	where we have denoted $\pi_s=\p(S_1=s)$ and $p_s=\p(g(X_1)=1, S_i=s), \ s=0,1,$ .
\end{theo}
\textbf{Proof:}

Consider for $i=1,\ldots,n,$ the random vectors
\begin{equation*}
Z_i=\left( \begin{array}{c}
\mathbbm{1}_{g(X_i)=1}\mathbbm{1}_{S_i=0}\\
\mathbbm{1}_{g(X_i)=1}\mathbbm{1}_{S_i=1}\\
\mathbbm{1}_{S_i=0}\\
\mathbbm{1}_{S_i=1}
\end{array} \right),
\end{equation*}
where $\mathbbm{1}_{g(X_i)=1}\mathbbm{1}_{S_i=s} \sim B(\p(g(X_i)=1,S_i=s))$ and $\mathbbm{1}_{S_i=s} \sim B(\p(S_i=s)), \ s=0,1,$. Thus, $Z_i$ has expectation
\begin{equation*}
\mathbb{E}Z_i=\left( \begin{array}{c}
\p(g(X_i)=1, S_i=0)\\
\p(g(X_i)=1, S_i=1)\\
\p(S_i=0)\\
\p(S_i=1)
\end{array} \right).
\end{equation*}
The elements of the covariance matrix $\Sigma_4$ of $Z_i$ are computed as follows:
\begin{align*}
Cov\left(\mathbbm{1}_{g(X_i)=1}\mathbbm{1}_{S_i=0}, \mathbbm{1}_{g(X_i)=1}\mathbbm{1}_{S_i=1}\right)= \mathbb{E}\left(\mathbbm{1}^2_{g(X_i)=1}\mathbbm{1}_{S_i=0}\mathbbm{1}_{S_i=1}\right)-\p(g(X_i)=1,S_i=0)\p(g(X_i)=1,S_i=1)
\end{align*}
\begin{align*}
Cov\left(\mathbbm{1}_{g(X_i)=1}\mathbbm{1}_{S_i=0}, \mathbbm{1}_{S_i=0}\right)&= \mathbb{E}\left(\mathbbm{1}_{g(X_i)=1}\mathbbm{1}^2_{S_i=0}\right)-\p(g(X_i)=1,S_i=0)\p(S_i=0)\\
&=\p(g(X_i)=1)\p(S_i=0)- \p(g(X_i)=1,S_i=0)\p(S_i=0)\\
&=\left[1- \p(S_i=0)\right]\p(g(X_i)=1,S_i=0)
\end{align*}
\begin{multline*}
Cov\left(\mathbbm{1}_{g(X_i)=1}\mathbbm{1}_{S_i=0}, \mathbbm{1}_{S_i=1}\right)=\mathbb{E}\left(\mathbbm{1}_{g(X_i)=1}\mathbbm{1}_{S_i=0}\mathbbm{1}_{S_i=1}\right)-\p(g(X_i)=1,S_i=0)\p(S_i=1)
\end{multline*}
\begin{multline*}
Cov\left(\mathbbm{1}_{g(X_i)=1}\mathbbm{1}_{S_i=1}, \mathbbm{1}_{S_i=0}\right)=\mathbb{E}\left(\mathbbm{1}_{g(X_i)=1}\mathbbm{1}_{S_i=0}\mathbbm{1}_{S_i=1}\right)-\p(g(X_i)=1,S_i=1)\p(S_i=0)
\end{multline*}
\begin{align*}
Cov\left(\mathbbm{1}_{g(X_i)=1}\mathbbm{1}_{S_i=1}, \mathbbm{1}_{S_i=1}\right)&= \mathbb{E}\left(\mathbbm{1}_{g(X_i)=1}\mathbbm{1}^2_{S_i=1}\right)-\p(S_i=1)\p(g(X_i)=1,S_i=1)\\
&=\p(g(X_i)=1,S_i=1)- \p(S_i=1)\p(g(X_i)=1,S_i=1)\\
&=\p(g(X_i)=1,S_i=1)\left[1- \p(S_i=1)\right]\\
&=\p(g(X_i)=1,S_i=1)\p(S_i=0)
\end{align*}
and finally,
\begin{equation*}
Cov(\mathbbm{1}_{S_i=0},\mathbbm{1}_{S_i=1})=\mathbb{E}\left(\mathbbm{1}_{S_i=0}\mathbbm{1}_{S_i=1}\right)-\p(S_i=0)\p(S_i=1)=-\p(S_i=0)\p(S_i=1).
\end{equation*}
From the Central Limit Theorem in dimension 4, we have that
\begin{equation*}
\sqrt{n}\left(\bar{Z}_n-\mathbb{E}Z_1\right) \xrightarrow{d} N_4\left(\mathbf{0}, \Sigma_4\right), \ as \ n\rightarrow \infty.
\end{equation*}
Now consider the function
\begin{center}
	$
	\begin{array}{cccc}
	\varphi:& \R^4 & \longrightarrow  &\R\\
	&(x_1,x_2,x_3,x_4) & \longmapsto & \displaystyle\frac{x_1x_4}{x_2x_3}
	\end{array}
	$
\end{center}
Applying the Delta-Method (see in~\cite{van1998asymptotic}) for the function $\varphi$, we conclude that
\begin{equation*}
\sqrt{n}\left(\varphi(\bar{Z}_n)-\varphi(\mathbb{E}Z_1)\right) \xrightarrow{d} \nabla\varphi^T\left(\mathbb{E}Z_1\right)N_4\left(\mathbf{0}, \Sigma_4\right), \ as \ n\rightarrow \infty,
\end{equation*}
where $\varphi(\bar{Z}_n)= T_n, \ \varphi(\mathbb{E}Z_1)=DI(g,X,S)$. \quad $\Box$

Hence, we can provide a confidence interval when estimating the disparate impact over a data set. Actually $\left(T_n \pm \frac{\sigma}{\sqrt{n}}Z_{1-\frac{\alpha}{2}}\right)$ is a confidence interval for the parameter $DI(g,X,S)$ asymptotically of level $1-\alpha$. \\ Previous theorem can be used to test the presence of disparate impact at a given level.
\begin{equation}
H_{0, \beta} : \ DI(g,X,S) \leqslant \beta \ \ \ vs. \ \ \ H_{1, \beta} : \ DI(g,X,S) > \beta
\end{equation}
aims at checking if $g$ has Disparate Impact at level $\beta$.
We want to check wether $ DI(g,X,S) \leq  \beta$.
Under $H_0$, the inequality $T_n-\beta \leqslant T_n -DI(g,X,S)$ holds, and so
\begin{equation*}
\frac{\sqrt{n}}{\sigma}\left(T_n-\beta\right) \leqslant \frac{\sqrt{n}}{\sigma} \left(T_n -DI(g,X,S)\right).
\end{equation*}
Finally, from the inequality above and Eq.~\eqref{eq:convergenceTn}, we have that
\begin{equation*}
\p_{H_0}\left(\frac{\sqrt{n}}{\sigma}\left(T_n- \beta \right) < Z_{1-\alpha} \right) 
\geqslant 
\p_{H_0}\left(\frac{\sqrt{n}}{\sigma}\left(T_n- DI(g,X,S) \right) < Z_{1-\alpha} \right) 
\longrightarrow 1- \alpha,
\end{equation*}
as $n \rightarrow\infty$ and, equivalently,
\begin{equation*}
\p_{H_0}\left(\frac{\sqrt{n}}{\sigma}\left(T_n-\beta\right) \geqslant Z_{1-\alpha} \right) \leqslant \p_{H_0}\left(\frac{\sqrt{n}}{\sigma}\left(T_n- DI(g,X,S) \right) \geqslant Z_{1-\alpha} \right) \longrightarrow \alpha,
\end{equation*}
as $n \rightarrow\infty$, where $Z_{1-\alpha}$ is the $(1-\alpha)$-quantile of $N(0,1)$. In conclusion, the test rejects $H_0$ at level $\alpha$ when
\begin{equation*}
\p_{H_0}\left(\frac{\sqrt{n}}{\sigma}\left(T_n-\beta\right) \geqslant Z_{1-\alpha} \right) \geqslant \alpha.
\end{equation*}

When dealing with Equality of Odds, we want to study the asymptotic behavior of the estimators of the True Positive and True Negative rates across both groups. The reasoning is similar for the two rates, so we will only show the convergence of the True Positive rate estimator, denoted in the following by $TP(g)$.

\begin{theo} \label{th:CE}
	Set the following estimate of the True Positive rate of a classifier $g$:
	\begin{equation*}
	R_n := \displaystyle\frac{\sum_{i=1}^{n}\mathbbm{1}_{g(X_i)=1}\mathbbm{1}_{Y_i=1}\mathbbm{1}_{S_i=0}\sum_{i=1}^{n}\mathbbm{1}_{Y_i=1}\mathbbm{1}_{S_i=1}}{\sum_{i=1}^{n}\mathbbm{1}_{g(X_i)=1}\mathbbm{1}_{g(X_i)=1}\mathbbm{1}_{S_i=1}\sum_{i=1}^{n}\mathbbm{1}_{Y_i=1}\mathbbm{1}_{S_i=0}}.
	\end{equation*}
	Then, the asymptotic distribution of this quantity is given by
	\begin{equation}\label{equ:convergenceTnCA1}
	\frac{\sqrt{n}}{\sigma}\left(R_n-TP(g)\right) \xrightarrow{d} N(0,1),  \ as \ n\rightarrow \infty,
	\end{equation}
	where $\sigma = \displaystyle\sqrt{\nabla\varphi^T\left(\mathbb{E}Z_1\right) \Sigma_4 \nabla\varphi\left(\mathbb{E}Z_1\right)}$
	and
	\begin{equation*}
	\nabla\varphi^T\left(\mathbb{E}Z_1\right)=\left(\frac{r_1}{p_1r_0},-\frac{p_0r_1}{p_1^2r_0}, -\frac{p_0r_1}{p_1r_0^2},\frac{p_0}{p_1r_0} \right)
	\end{equation*}
	\begin{equation*}
	\Sigma_4= \left(\begin{array}{cccc}
	p_0(1-p_0) & & & \\
	-p_0r_1 & p_1(1-p_1) & & \\
	p_0(1-r_0) & -p_1r_0 & r_0(1-r_0) & \\
	p_0r_1 & p_1(1-r_1) & -r_0r_1 & r_1(1-r_1)
	\end{array} \right),
	\end{equation*}
	where we have denoted $p_s=\p(g(X_1)=1, Y_1=1, S_1=s),$ and $r_s=\p(Y_1=1,S_1=s),$ for $s=0,1.$
\end{theo}

Proof of Theorem~\ref{th:CE}
The proof follows the same guidelines of previous proof. We set here
\begin{equation*}
Z_i=\left( \begin{array}{c}
\mathbbm{1}_{g(X_i)=1}\mathbbm{1}_{Y_i=1}\mathbbm{1}_{S_i=0}\\
\mathbbm{1}_{g(X_i)=1}\mathbbm{1}_{Y_i=1}\mathbbm{1}_{S_i=1}\\
\mathbbm{1}_{Y_i=1}\mathbbm{1}_{S_i=0}\\
\mathbbm{1}_{Y_i=1}\mathbbm{1}_{S_i=1}
\end{array} \right),
\end{equation*}
where $\mathbbm{1}_{g(X_i)=1}\mathbbm{1}_{Y_i=1}\mathbbm{1}_{S_i=s} \sim B(\p(g(X_i)=1,Y_i=1,S_i=s))$ and $\mathbbm{1}_{Y_i=1}\mathbbm{1}_{S_i=s} \sim B(\p(Y_i=1,S_i=s)), \ s=0,1,$. 
From the Central Limit Theorem, we have that
\begin{equation*}
\sqrt{n}\left(\bar{Z}_n-\mathbb{E}Z_1\right) \xrightarrow{d} N_4\left(\mathbf{0}, \Sigma_4\right), \ as \ n\rightarrow \infty.
\end{equation*}
with 
\begin{equation}
\Sigma_4= \left(\begin{array}{cccc}
p_0(1-p_0) & & & \\
-p_0r_1 & p_1(1-p_1) & & \\
p_0(1-r_0) & -p_1r_0 & r_0(1-r_0) & \\
p_0r_1 & p_1(1-r_1) & -r_0r_1 & r_1(1-r_1)
\end{array} \right).
\end{equation}
Now consider the function
\begin{center}
	$
	\begin{array}{cccc}
	\varphi:& \R^4 & \longrightarrow  &\R\\
	&(x_1,x_2,x_3,x_4) & \longmapsto & \displaystyle\frac{x_1x_4}{x_2x_3}
	\end{array}
	$
\end{center}
Applying the Delta-Method for the function $\varphi$, we conclude that
\begin{equation*}
\sqrt{n}\left(\varphi(\bar{Z}_n)-\varphi(\mathbb{E}Z_1)\right) \xrightarrow{d} \nabla\varphi^T\left(\mathbb{E}Z_1\right)N_4\left(\mathbf{0}, \Sigma_4\right), \ as \ n\rightarrow \infty,
\end{equation*}
where $\varphi(\bar{Z}_n)= R_n,$ and $\varphi(\mathbb{E}Z_1)=TP(g)$.
\quad $\Box$

\subsection{Bootstraping v.s Direct Calculation of IC interval} \label{s:boot}

\begin{figure}[h]
	\centering
	\includegraphics[scale=0.6]{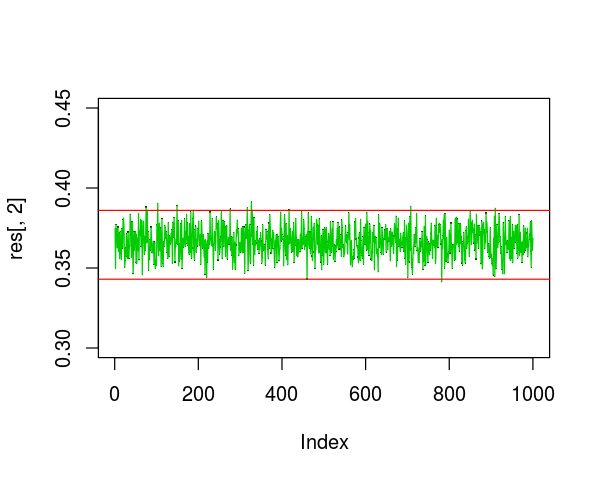}
	\caption{Comparison with bootstrap computations}
	\label{fig:bootIC}
\end{figure}

The estimation of the Disparate Impact is unstable. In this paper we promote the use of the theoretical confidence interval based on the well known Delta method to control its variability. Contrary to \cite{doi:10.1111/j.1744-6570.2000.tb00195.x}, it does not rely on Gaussian approximation. We compare the stability of this confidence interval to bootstrap simulations, see for instance in \cite{efron1994introduction} for more details on bootstrap methods. \\
\indent  For this we build 1000 bootstrap replicates and estimate the disparate impact. Figure~\ref{fig:bootIC} presents the simulations. We can see that the bootstrap simulations remain in the confidence interval. Moreover, if we build a confidence interval for the bootstrap estimator, the confidence intervals are the same. We obtain by the theoretical confidence interval $[0.349, 0.384]$ while the bootstrap's confidence interval is $[0.349, 0.385]$. \vskip .1in
 Hence the theoretical confidence is a reliable measure of fairness for the data set and should be preferred due to its small computation time compared to the 1000 bootstrap replication. \\
 Note that in this paper, for sake of clarity, we have chosen to focus only on the disparate impact criterion. Yet all other fairness criteria should be given with the calculation of a confidence interval. For instance in \cite{del2019central} we propose confidence intervals for Wasserstein distance which is use in many methods in fair learning.

\subsection{Application to other real datasets} \label{s:use}
To illustrate these tests we have also considered another two well-known and real data sets.
\begin{enumerate} 
	
	\item \textbf{\textit{German Credit} data.}
	This data set is often claimed to exhibit some origin discrimination in the success of being given a credit  by the German bank. Hence we compute the disparate impact w.r.t Origin. We obtain
	$$ DI = 0.77 \in [0.68 , 0.87 ] .$$
	Hence here confidence intervals play an important role. Actually the disparate impact is not statistically significantly lower than 0.8, which entails that the discrimination of the decision rule of the German bank can not be shown, which promotes the use of a proper confidence interval.   \vskip .1in
	\item \textbf{\textit{COMPAS Recidivism data}}. A third  data set is composed by the data of the controversial COMPAS score detailed in \cite{dieterich2016compas}.  The data is composed of 7214 offenders with personal variables observed over two years. A score predicts their level of dangerosity which determines whether they can be released while a variable points out if there has been recidivism. Hence Recidivism of offenders is predicted using a score and confronted to possible racial discrimination which corresponds to the protected attribute. The protected variable separates the population into caucasian and non caucasian.  To evaluate the level of discrimination we first compute the disparate impact with respect to the true variable and the COMPAS score seen as a predictor. 
	$$ DI= 0.76 \in [.72,.81]; \quad DI({\rm COMPAS})= 0.71 \in [0.68;0.74].$$ 
	In both cases, the data are biased but the level of discrimination is low. Yet as mentioned in al the studies on this data set, the level of errors of prediction is significantly different according to the ethnic origin of the defender. Actually the conditional accuracy scores and their corresponding confidence intervals show clearly the unbalance treatment received by both populations.
	$$
	TPR= 0.6 \in [ 0.54,	 	0.65 ]$$
	$$TNR=3.38 \in [2.46,4.3 ]$$
	This unbalanced treatment is clearly assessed with the confidence interval.
\end{enumerate}

\end{document}